\pdfoutput=1 
\documentclass[11pt]{article}
\usepackage[]{EMNLP2023}

\usepackage{times}
\usepackage{latexsym}
\usepackage[T1]{fontenc}
\usepackage[utf8]{inputenc}
\usepackage{microtype}
\usepackage{inconsolata}
\usepackage{graphicx}
\usepackage{multirow}
\usepackage{booktabs}
\usepackage{tipa}
\usepackage{paralist}
\usepackage[inline]{enumitem}
\usepackage{subcaption}
\usepackage{caption}
\usepackage{adjustbox}
\usepackage{array}
\usepackage{blindtext}
\usepackage{hyperref}
\usepackage{siunitx}
\usepackage{fdsymbol}
\newcolumntype{R}[2]{%
    >{\adjustbox{angle=#1,lap=\width-(#2)}\bgroup}%
    l%
    <{\egroup}%
}

\title{PyThaiNLP: Thai Natural Language Processing in Python}

\author{
  Wannaphong~Phatthiyaphaibun\textsuperscript{$\vardiamondsuit$},
  Korakot~Chaovavanich\textsuperscript{\dagger},
  Charin~Polpanumas\textsuperscript{\dagger}, \\
  \textbf{Arthit~Suriyawongkul\textsuperscript{\ddagger},
  Lalita~Lowphansirikul\textsuperscript{$\vardiamondsuit$},
  Pattarawat~Chormai\textsuperscript{\mathsection \mathparagraph},} \\ 
  \textbf{Peerat~Limkonchotiwat\textsuperscript{$\vardiamondsuit$},
  Thanathip~Suntorntip\textsuperscript{$\clubsuit$},
  Can~Udomcharoenchaikit\textsuperscript{$\vardiamondsuit$}} \\
  \textsuperscript{$\vardiamondsuit$}VISTEC,
  \textsuperscript{\dagger}PyThaiNLP,
  \textsuperscript{\ddagger}Trinity College Dublin, \\
  \textsuperscript{\mathsection}Technische Universit\"at Berlin,
  \textsuperscript{\mathparagraph}Max Planck School of Cognition,
  \textsuperscript{$\clubsuit$}Wisesight\\
  \texttt{wannaphong.p\_s21@vistec.ac.th}
}

\begin{document}
\maketitle
\begin{abstract}

We present PyThaiNLP, a free and open-source natural language processing (NLP) library for Thai language implemented in Python. It provides a wide range of software, models, and datasets for Thai language. We first provide a brief historical context of tools for Thai language prior to the development of PyThaiNLP. We then outline the functionalities it provided as well as datasets and pre-trained language models. We later summarize its development milestones and discuss our experience during its development. We conclude by demonstrating how industrial and research communities utilize PyThaiNLP in their work. The library is freely available at \url{https://github.com/pythainlp/pythainlp}.

\end{abstract}

\section{Introduction}

In recent years, the field of natural language processing has witnessed remarkable advancements, catalyzing breakthroughs for various applications. However, Thai has remained comparatively underserved due to the challenges posed by limited language resources \citep{arreerard-etal-2022-survey}.

Thai is the de facto national language of Thailand. It belongs to Tai linguistic group within the Kra-Dai language family. According to Ethnologue \citep{ethnologue}, there are 60.2 million users of Central Thai, of which 20.8 million are native (2000). If including the Northern (6 million, 2004), Northeastern (15 million, 1983), and Southern (4.5 million, 2006) variants, there are estimated 85.7 million users of Thais speakers around the world.

Thai is a scriptio continua or has neither spaces nor other marks between the words or sentences in its most common writing style \citep{sornlertlamvanich-etal-2000}. The lack of clear word and sentence boundaries leads to ambiguity that cannot be disambiguated using merely just grammatical knowledge \citep{supnithi-etal-2004}.

Although many closed-source open APIs for NLP have an ability to process Thai language\footnote{Such as those provided by commercial cloud service providers and ``AI for Thai'', the government-funded Thai AI service platform at \url{https://aiforthai.in.th/}.}, we believe that an open-source toolbox is essential for both researchers and practitioners to not only access the NLP capabilities but also gain full transparency and trust on both training data and algorithms.\footnote{For a discussion about concentrated power and the political economy of the `open' AI, see \citet{widder2023business}.} This allows the community to adapt and further develop the functionalities as needed, making a crucial step towards democratizing NLP.

This paper introduces PyThaiNLP, an open-source Thai natural language processing library written in Python programming language. Its features span from a simple dictionary-based word tokenizer, to a statistical named-entity recognition, and an instruction-following large language model. The library was released in 2016 under an Open Source Initiative-approved Apache License 2.0 that allows free use and modification of software, including commercial use.

\section{Open-source Thai NLP before PyThaiNLP}

Before PyThaiNLP started in 2016, some free and open-source software do exist for different Thai NLP tasks, but there were no unified open-source toolkits that unified multiple tools or tasks in a single library, and the number of available Thai NLP datasets was low compared to high-resource languages like Chinese, English, or German.

Natural Language Toolkit (NLTK) \citep{bird-loper-2004-nltk}, one of the most comprehensive and most popular NLP libraries in Python at the time, did not support Thai. OpenNLP, another popular free and open-source NLP toolkit written in Java, started having Thai models in version 1.4 (2008)\footnote{\url{https://opennlp.sourceforge.net/models-1.4}. Its README from December 2008 also mentioned Thai components: \url{https://web.archive.org/web/20081219153426/http://opennlp.sourceforge.net/README.html} } but in version 1.5 (2010) Thai was no longer listed in its supported languages\footnote{\url{https://opennlp.sourceforge.net/models-1.5}. \citet{arreerard-etal-2022-survey}, however, reports that Apache OpenNLP supports these basic Thai NLP tasks: word tokenization, part-of-speech tagging, and sentence detection.}.

Open Thai language resources, like annotated corpora, were also limited in size and number. ``Publicly available'' datasets tend to have restricted access, either through restrictive licenses\footnote{ Even today, this practice continues: take, for instance, the LST20 corpus from NECTEC, which has multiple layers of linguistic annotation. However, the free version can only be used for non-commercial purposes. See \url{ https://opend-portal.nectec.or.th/en/dataset/lst20-corpus}. } or the registration requirement, or both.

Because there is a few toolkits available, limited in documentation and performance, short of rigorous benchmarking, and/or lack of maintenance, Thai NLP reseachers had to spend their limited time and resources building basic components and/or collecting a dataset before they could proceed further for more advanced problems. The limited availability of source codes and datasets also affects reproducibility.

Examples of Thai NLP tools and datasets before PyThaiNLP (pre-2016):
\begin{itemize}
    \item \textbf{Word tokenization:} ICU BreakIterator \citep{icu1999} [Unicode License] based on \citet{gillam1999text}, LibThai \citep{libthai2001} [LGPL], KU Wordcut \citep{sudprasert2003thai} [GPL], SWATH \citep{charoenpornsawat2003swath} [GPL] based on \citet{meknavin_feature-based_1997}, LexTo \citep{nectec2006lexto} [LGPL], OpenNLP \citep{bierner2007opennlp} [LGPL], TLex \citep{haruechaiyasak2009tlex} [Freeware], and wordcutpy \citep{wordcutpy2015} [LGPL]. \citet{choochartetal2008} provided a comparative study of some of these tools.
    \item \textbf{Part-of-speech (POS) tagging:} OpenNLP and RDRPOSTagger \citep{nguyen-etal-2014-rdrpostagger} [GPL] support Thai POS tagging. There are corpora such as ORCHID \citep{sornlertlamvanich1999building} and NAiST \citep{kawtrakul-etal-2002-state} which provide not only POS but also word boundaries.
    \item \textbf{Named-entity recognition (NER):} Polyglot \citep{polyglot2015} [GPL], a multilingual NLP software, supports Thai NER based on \citet{polyglotner2015}. For datasets, BEST-2009 corpus \citep{kosawat2009best} is available but cannot be used commercially, as its license is Creative Commons Attribution-NonCommercial-ShareAlike Public License.
    \item \textbf{Automatic speech recognition (ASR):} Thai Language Audio Resource Center (Thai\hspace{0pt}ARC) corpus \citep{hoonchamlong_thai_1997} provides audio recordings of dialects and speech styles, with transcripts; it is not designed specifically for ASR. NECTEC-ATR \citep{nectec-atr}, LOTUS \citep{lotus_asr}, LOTUS-BN \citep{LOTUS-BN}, LOTUS-Cell \citep{chotimongkol2010development}, CU-MFEC \citep{cumfec_corpus} and TSync-2 are ASR corpora for different domains and tasks; their licenses are not fully open. See \citet{charoenporn-etal-2004-open}, \citet{wutiwiwatchai2007}, and \citet{cumfec_corpus} for reviews.
\end{itemize}

Apart from the ones listed above, more open-source Thai word tokenizers were released after 2009 as a result of BEST (Benchmark for Enhancing the Standard of Thai language processing) evaluation for Thai word segmentation organized by the National Electronics and Computer Technology Center (NECTEC) in 2009 \citep{kosawat2009interbest}, and 2010\footnote{\url{https://thailang.nectec.or.th/archive/indexa290.html}}. Unfortunately, these tokenizers are no longer maintained and are not accessible at the time of writing. The most impactful contribution from BEST, however, is the BEST-2010 word segmentation dataset that was publicly released. This dataset provides a basis for a lot of modern Thai open-source word segmentation software.

We should also mention the Thai Language Toolkit (TLTK) \citep{tltk}. Its first release on Python Package Index (version 0.3.4, February 2018) includes statistical syllable and word segmentation \citep{wirote2002}, POS tagging, and spelling suggestion. Its latest version, as of writing, features discourse unit segmentation, NER, grapheme-to-phoneme conversion, IPA transcription, romanization, and more. To date, TLTK and PyThaiNLP are the only two comprehensive Thai NLP libraries for Python. However, TLTK's documentation is still quite limited. For more reviews on Thai NLP tools and datasets, including more recent ones (post-2016), see \citet{arreerard-etal-2022-survey}.

\section{PyThaiNLP and Its Ecosystem}

\begin{figure*}
    \centering
    \includegraphics[width=\linewidth]{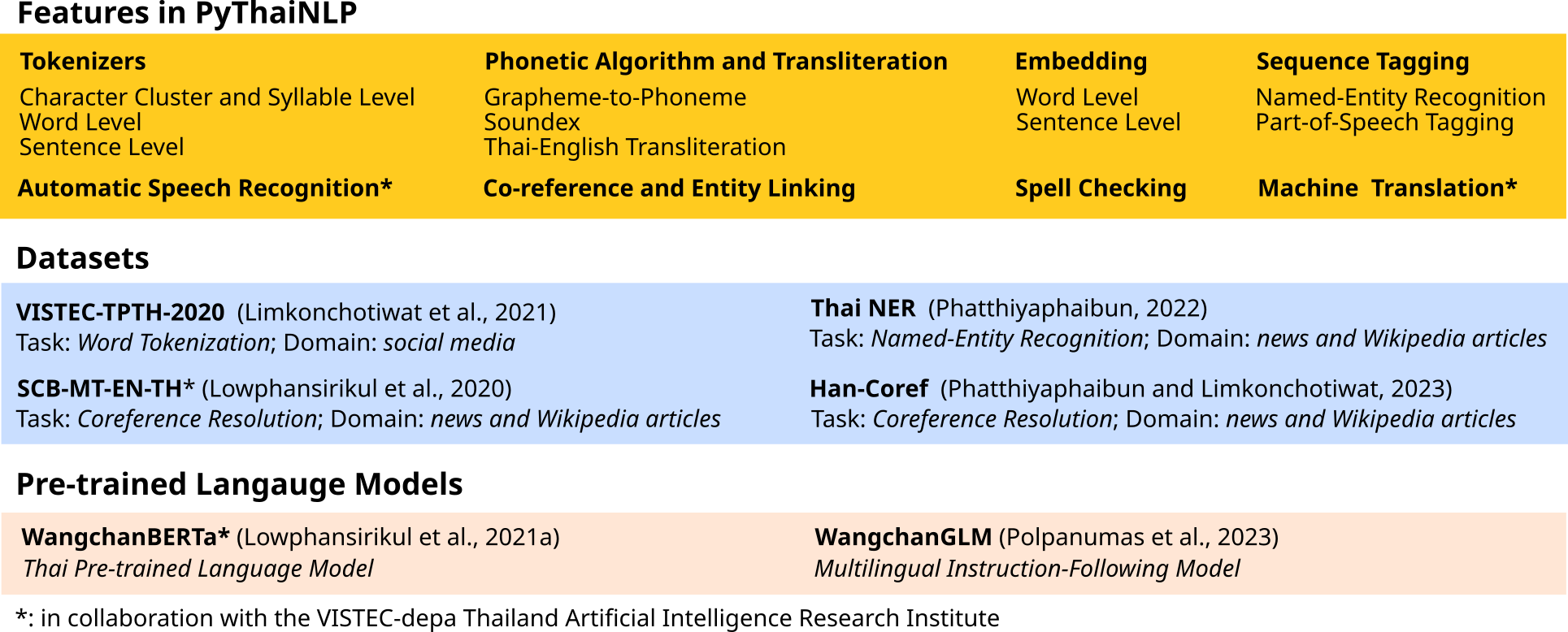}
    \caption{Functionalities, datasets, and pre-trained language models available in PyThaiNLP's ecosystem.}
    \label{fig:ecosystem}
\end{figure*}

Our primary objective is to ensure the user-friendliness and simplicity of the library. Drawing inspiration from NLTK, we follow numerous established interfaces. For example, \texttt{word\_tokenize} and \texttt{pos\_tag}. In addition, we also create datasets and pre-trained models for the Thai language. Figure \ref{fig:ecosystem} illustrates the overview of PyThaiNLP's functionalities and its ecosystem. Table \ref{table:timeline} displays the development milestones of PyThaiNLP.

We will discuss here only popular features and major datasets/models.

\subsection{Features}

\subsubsection{Word and Sentence Tokenization}

PyThaiNLP supports many word tokenization algorithms.\footnote{For the ease of experimenting with different word tokenization algorithms, Pattarawat Chormai has created a Thai word tokenizers collection as a Docker container image: \url{https://github.com/PyThaiNLP/docker-thai-tokenizers}.} The default algorithm is NewMM\, which is dictionary-based maximum matching \citep{virach1993wordseg} and utilizes Thai character cluster \citep{theeramunkong2000character}. The pure-Python tokenizer performs reasonably well on public benchmarks. \citet{chormai-etal-2020-syllable} demonstrated that it is the fastest word tokenizer on the BEST 2010 benchmark, with \SI{71.18}{\percent} accuracy (compared to state-of-the-art at \SI{95.60}{\percent}). Thanathip Suntorntip ported NewMM to Rust programming language\footnote{\url{https://github.com/pythainlp/nlpo3}}, resulting in an even faster word tokenizer in our toolbox.

For sentence tokenization, we trained a conditional random field (CRF) model, using python-crfsuite \citep{pythoncrfsuite}, on translated TED transcripts and Thai sentence boundaries are assumed to be denoted by English sentence boundaries \citep{Lowphansirikul_2021}.

\subsubsection{Spell Checking}

For spell checking, we have many engines; the \citet{norvig2007write} one uses a spelling dictionary from Thai National Corpus \citep{aroonmanakun2009thai}, symspellpy \citep{symspellpy} that is a Python port of SymSpell v6.7.1, and phunspell \citep{phunspell} that is a port of Hunspell.

\subsubsection{Phonetic Algorithm and Transliteration}

PyThaiNLP supports a couple of grapheme-to-phoneme (g2p) conversion engines. We trained Thai-g2p model with data from Wiktionary\footnote{\url{https://www.wiktionary.org/}}, a free online dictionary.

PyThaiNLP implemented many Thai Soundex algorithms. For example, \citet{lorchirachoonkul1982thai}, \citet{udompanich1983}, Thai-English cross-language Soundex \citep{suwanvisat1998thai}, and MetaSound (Metaphone-Soundex combination) \citep{Snae2009NovelPN}.

PyThaiNLP supports the following transliteration implementations: Thai romanization using the Royal Thai General System of Transcription (RTGS), transliteration of romanized Japanese/Korean/Mandarin/Vietnamese texts to Thai using Wunsen library \cite{wunsen2022}\footnote{The library implements various transliteration systems that recommended by the Royal Society of Thailand.}, and Thai word pronunciation.

\subsubsection{Sequence Tagging (NER and POS)}

We create a named-entity recognition model called Thai NER \citep{wannaphong_phatthiyaphaibun_2022_7761354} by finetuning the WangchanBERTa model~\citep{lowphansirikul2021wangchanberta} and CRF model.

For part-of-speech tagging, we trained a CRF tagger, a perceptron tagger \citep{pos200}, a unigram tagger, and finetuned the WangchanBERTa model. The POS training sets are derived from ORCHID corpus \citep{sornlertlamvanich1999building}, Blackboard Treebank annotated based on the LST20 Annotation Guideline \citep{boonkwan2020annotation}, and Parallel Universal Dependencies (PUD) treebanks \citep{smith-etal-2018-82}.

\subsubsection{Coreference Resolution and Entity Linking}

For coreference resolution, we create Han-Coref, a Thai coreference resolution corpus and model \citep{wannaphong_phatthiyaphaibun_2023_7965488}.

For entity linking, PyThaiNLP supports it using BELA model \citep{plekhanov2023multilingual}.

\subsubsection{Word Embeddings}

We extract token embeddings from our thai2fit \citep{charin_polpanumas_2021_4429691}, a word-level ULMFiT language model \citep{howard-ruder-2018-universal} \cite{howard2020fastai} trained on Thai Wikipedia, and use them as word embeddings for PyThaiNLP. It was the state-of-the-art pre-trained model in many Thai classification benchmarks \citep{cstorm125_2020_3852912} before the multilingual BERT model was released \citep{pycon_youtube}.

\subsubsection{Machine Translation}

We collaborated with VISTEC-depa Thailand Artificial Intelligence Research Institute (AIResearch.in.th)\footnote{ AIResearch.in.th is an initiative co-funded by a research university and a government agency, namely Vidyasirimedhi Institute of Science and Technology (VISTEC) in Wang Chan, Rayong, and the Digital Economy Promotion Agency (depa) under the Ministry of Digital Economy and Society, to create AI infrastructure for Thailand. } to create the English-Thai translation dataset and model. The model outperformed Google Translate on an out-of-sample test set at the time of release \citep{Lowphansirikul_2021}.

\subsubsection{Automatic Speech Recognition}

In order to develop a dataset for ASR, PyThaiNLP members contribute to the development of Common Voice corpus \citep{ardila-etal-2020-common}, including Thai sentence cleanup and validation rules for its Sentence Collector\footnote{\url{ https://github.com/common-voice/sentence-collector }}, an online campaign inviting people to contribute Thai sentences, and offline events for volunteers to contribute their voices and voice validation.

Utilizing Common Voice Corpus 7.0, we created a Thai ASR model in collaboration with AIResearch.in.th and achieved the lowest character error rate in a benchmark \citep{vistec-depa_ai_research_institute_of_thailand_2023}.

\subsection{Datasets}

\subsubsection{VISTEC-TPTH-2020: Word Tokenization, Spell Checking and Correction}

VISTEC-TPTH-2020 is a Thai word tokenization and spell checking dataset in the social media domain, the largest one to date \citep{limkonchotiwat-etal-2021-handling}. We collected 50,000 sentences from top trending posts on Twitter in 2020 and selected only posts with substantial character counts. This dataset is a multi-task dataset, including mention detection, spell checking, and spell correction.

\subsubsection{Thai NER: Named Entity Recognition}

Thai NER is a Thai named-entity recognition dataset. We curated text from various domains including news, Wikipedia articles, government documents, as well as text from other Thai NER datasets. The data is manually re-labeled for consistency \citep{wannaphong_phatthiyaphaibun_2022_7761354}.

\subsubsection{Han-Coref: Coreference Resolution}

Han-Coref is a coreference resolution dataset containing 1,339 documents in news and Wikipedia domains \citep{wannaphong_phatthiyaphaibun_2023_7965488}.

\subsubsection{scb-mt-en-th-2020: English-Thai Machine Translation}

scb-mt-en-th-2020 is an English-Thai sentence pair dataset consisting of 1,001,752 text pairs \citep{Lowphansirikul_2021}. It is a collaborative work with AIResearch.in.th.

\subsection{Pre-trained Language Models}

WangchanBERTa is an encoder-only pre-trained Thai language model. Based on public benchmarks, it is the current state-of-the-art \citep{lowphansirikul2021wangchanberta}. It is also a collaborative work with AIResearch.in.th.

WangChanGLM \citep{charin_polpanumas_2023_7878101} is a multilingual instruction-following model finetuned from XGLM \citep{lin-etal-2022-shot}.

\section{Community and Project Milestones} \label{sec:milestones}

\begin{table*}[!ht]
\centering
\begin{tabular}{|l|l|}
\hline
\textbf{Years} & \textbf{Notable Features} \\ \hline
2016           & Word tokenization, part-of-speech tagging \\ \hline
2017           & Soundex, spell checking, WordNet support \\ \hline
2018           & Text classification language model, NER corpus/model, date and time parsing/formatting \\ \hline
2019           & Syllable tokenization, date and time spell out \\ \hline
2020           & ASR model, machine translation dataset/model, grapheme-to-phoneme conversion \\ \hline
2021           & Autoencoding language model, word-to-phoneme conversion \\ \hline
2022           & Dependency parsing, nested NER, text augmentation \\ \hline
2023           & Coreference resolution dataset/model, generative language model \\ \hline
\end{tabular}
\caption{\textcolor{black}{Notable features introduced to PyThaiNLP over the years.}}
\label{table:timeline}
\end{table*}

\subsection{Foundation Years (2016-2019)}

Wannaphong Phatthiyaphaibun, a high school student at the time, created PyThaiNLP in 2016 as a hobby project. He wanted to create a simple Thai chatbot in Python. He used PyICU as a word tokenizer and soon found out that Thai language did not have a comprehensive NLP toolkit in Python like NLTK \citep{bird-loper-2004-nltk}. He decided to create PyThaiNLP and hosted the project on GitHub\footnote{\url{https://github.com/pythainlp/pythainlp}}.

After the first few official releases, following Korakot Chaovavanich's suggestion, a ``Thai Natural Language Processing'' group has been created as a public Facebook group\footnote {\url{https://www.facebook.com/groups/thainlp}}. This serves as a main venue to showcase PyThaiNLP's capabilities and a hub for Thai NLP researchers and practitioners to discuss the field. Today, the group has over 16,000 members and is Thailand's largest NLP interest group. This communication channel also performs a recruiting function for us. The first offline meetup of the group occurred in 24 May 2018 as a bird-of-a-feather session after a Data Science BKK meetup\footnote{\url{https://www.facebook.com/groups/thainlp/permalink/564348637279964/}}.

\begin{figure}
    \centering
    \includegraphics[clip,trim=0.23cm 0.8cm 0.23cm 0, width=\linewidth]{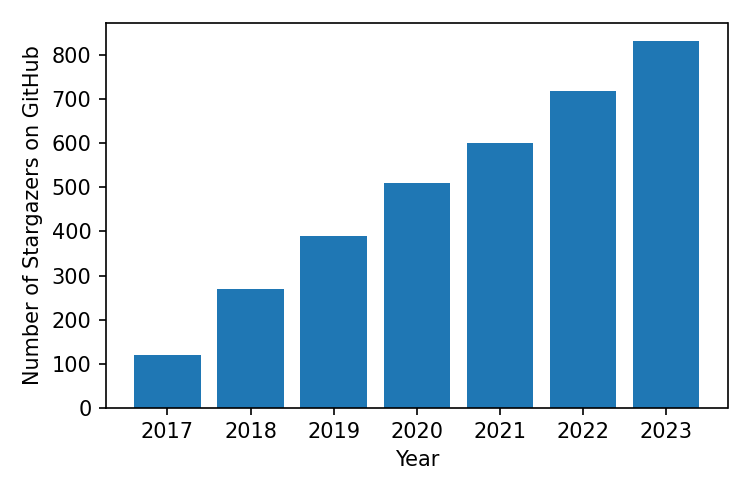}
    \caption{Number of stars PyThaiNLP has received from GitHub users over the years.}
    \label{fig:star}
\end{figure}

Many of our main contributors, such as Charin Polpanumas and Arthit Suriyawongkul organically joined the project from the community. At this stage, we created foundational capabilities such as word tokenization, part-of-speech tagging, subword tokenization, named-entity recognition, and word vectors. A lot of code cleaning, reorganization, and documentation also happened around 2018-2019. This included the adoption of PEP 484 type hints\footnote{\url{https://peps.python.org/pep-0484/}} and other Python best practices to make the code even more readable and facilitate off-line type checkers. The adoption of PyThaiNLP can be reflected by the number of stars on GitHub the project received over the years (Figure \ref{fig:star}).

\subsection{Gaining Resources for Large Language Models (2019-present)}

The growing activity of PyThaiNLP development can be seen from the number of code commits to the Git repository, which reached its peak in Q4 2019\footnote{\url{https://github.com/PyThaiNLP/pythainlp/graphs/contributors}}. In 2020, the project began a collaboration with AIResearch.in.th. Their main focus was to create and distribute open-source models and datasets. This collaboration has provided PyThaiNLP with computational resources we need to scale up our operations as well as additional developers for maintaining the project, such as Lalita Lowphansirikul.

Under the collaboration, we have built an English-Thai sentence pair dataset and the state-of-the-art English-Thai translation model \citep{Lowphansirikul_2021}, the RoBERTa-based monolingual language model WangchanBERTa \citep{lowphansirikul2021wangchanberta}, and most recently the multilingual instruction-following model WangChanGLM \citep{charin_polpanumas_2023_7878101}.

Due to limited computational and human resources, we prioritize features with the highest impact-to-effort ratio. For example, during 2019-2020, there were two types of dominant transformer-based language models: encoder-only BERT family and decoder-only GPT family. We opted to pursue the encoder-only models and trained WangchanBERTa because, at the time, it required relatively fewer resources to train and had better performance across impactful tasks such as text classification, sequence tagging, and extractive question answering. It was not until decoder-only models proved to create more value-added in 2022 that we started to train such models as WangChanGLM.

\subsection{Community and Infrastructure for Software Quality} \label{sec:software-quality}

It is important to be noted that the community not only made contributions in the form of feature improvements but also in the areas of documentation, including computational documentation (e.g., Jupyter notebooks), improving code quality and test suite, and streamlining software testing and delivery. Some of which may not be visible to the users but are crucial for the development of the project.

On the infrastructure side, test automation and continuous integration (CI) helps us systematically reinforce code style, detect code security vulnerabilities, maintain code coverage, and test the library in different computer configurations.

We were since 2017 rely on free Travis CI\footnote{\url{https://www.travis-ci.com/}} and AppVeyor\footnote{\url{https://www.appveyor.com/}} for continuous integration workflow and later in June 2020 completely migrated to GitHub Actions\footnote{\url{https://github.com/features/actions}}. Every GitHub pull requests will go through Black\footnote{\url{https://github.com/psf/black}} for code formatting and Flake8\footnote{\url{https://flake8.pycqa.org}} for PEP 8 code style\footnote{\url{https://peps.python.org/pep-0008/}} and cyclomatic complexity checks \citep{mccabe-1976-complexity}. pip installation package will be built and tested against the test suite in Linux, macOS, and Windows\footnote{ Easy installation and consistent behavior across platforms are what we aim for. This is one of the reasons why we developed a pure-Python NewMM. The previous implementation of our default word tokenizer requires marisa-trie, a trie data structure library in C++. Unfortunately, marisa-trie does not officially support mingw32 compiler on Windows. }. The package then can be automatically publish to the Python Package Index directly from the CI, once it passed all the tests in every platform. 

PyThaiNLP code coverage reached \SI{80}{\percent} towards the end of 2018, compare to under \SI{60}{\percent} in 2017. Code coverage is a metric that can help assess the quality of the test suite, and it therefore reflects how well the functionalities are thoroughly tested. The coverage went over \SI{90}{\percent} in August 2019 and kept stable at this level until 2022\footnote{ Our code coverage is measured by coverage.py which is included in our continuous integration workflow. The coverage stats are made available online by Coveralls at: \url{https://coveralls.io/github/PyThaiNLP/pythainlp} }.

From early 2022, we experienced a gradual drop of the code coverage to \SI{80}{\percent}. The main reason is a growing number of features that require a large language model that cannot fit inside our standard GitHub-hosted runners. We have to remove some of the tests for those features. Before 2022, we also tested our library against versions of CPython and PyPy, but now it has been reduced to only CPython 3.8 due to the lack of support for other Python versions in some of our machine learning dependencies.

Some of the common code improvements we made after analyzing code coverage and other tests were the removal of unused code, fixing inconsistent behavior in different operating systems, better handling of a very long string, empty string, empty list, null, and/or negative values, and better handling of exceptions in control flow, resulting a code that is smaller and more robust.

\section{PyThaiNLP in the Wild}

\subsection{PyThaiNLP and Its Research Impact}

Researchers worldwide use PyThaiNLP to work with Thai language. For instance, for word tokenization in cross-lingual language model pretraining \citep{lample2019cross}, universal dependency parsing \citep{smith-etal-2018-82}, and cross-lingual representation learning \citep{conneau-etal-2020-unsupervised}. In addition, research and industry-grade tools namely SEACoreNLP\footnote{\url{https://seacorenlp.aisingapore.net/docs/}}, an open-source initiative by NLPHub of AI Singapore, and spaCy \cite{Honnibal_spaCy_Industrial-strength_Natural_2020} include PyThaiNLP as part of their toolkit.

\subsection{PyThaiNLP and Its Industry Impact}

PyThaiNLP is used in many real-world business use cases in firms of all sizes both domestic and international. User feedback generally highlights how the library has sped up their product development cycles involving Thai NLP as well as its effectiveness in terms of business outcomes. The most frequently used functionalities are tokenization and text normalization. We introduce here selected use cases from national and multinational firms in banking, telecommunication, insurance, retail, and software development.

\textbf{Siam Commercial Bank (BKK:SCB; USD 10B market cap)} is one of Thailand's largest banks. The bank operates a chatbot to automatically answer customer queries. Their data analytics team finetuned WangchanBERTa for intent classification to enhance its question-answering capabilities as well as to detect personal information in customers' inputs in order to exclude them from their internal training sets. Moreover, the team relies on basic text processing functions such as tokenization and normalization to speed up their development process. They have also found the published performance benchmarks to be useful when selecting models for their tasks.

\textbf{True Corporation (BKK:TRUE; 6B)} is one of the two providers in Thailand's duopoly telecommunication market. Its subsidiary, True Digital Group, uses PyThaiNLP both for digital media analysis and for recommendation engine on production. They featurized their Thai-text contents using thai2fit word vectors and saw a noticeable uplift in user engagement and subsequent business outcomes. They also combined our word vectors with Top2Vec \citep{angelov2020top2vec} to perform topic modeling and improve customer experience.

\textbf{Central Retail Digital (BKK:CRC; 6B)} is a digital transformation unit serving Central Retail, Thailand's largest department store. Their data science team used PyThaiNLP mainly to enhance search and recommendation offerings across five business units and other six million customers. Word tokenization and text normalization were used to preprocess product information and search queries as input for the product search system. Since most search systems are built for languages with white spaces as word delimiters, this preprocessing step has allowed their product search to outperform out-of-the-box solutions which are not compatible with Thai. For content-based recommendations, the team featurized production information to create a model that recommends similar products to customers.

\textbf{AIA Thailand (HKG:1299; 109B global)} is the Thai headquarter of the global insurance firm American Insurance Association. Their data science team employs PyThaiNLP in analyzing their inbound and outbound call logs using word tokenization, text normalization, stop word handling, and local-time-format string handling functionalities. For the inbound calls, they normalize and tokenize the logs to perform topic modeling and identify critical topics of conversation to emphasize both automated voice bot and human staff training and allocation. This resulted in improved percentage of calls that the voice bot fulfilled successfully and reduced call waiting time. For the outbound calls, they perform keyword identification from the logs processed by PyThaiNLP to gain insights to improve customer retention.

\textbf{VISAI} is a VISTEC university spin-off that provides machine learning tools and consulting services. It has finetuned WangchanBERTa to perform text classification, named entity recognition, and relation extraction on unstructured data of their clients to create a queryable knowledge graph. They also use tokenization and text normalization functionalities to facilitate text processing for all their NLP-based products.

\section{Conclusion and Future Works}

This paper introduces the PyThaiNLP library, explains its features and datasets (as illustrated in Figure \ref{fig:ecosystem}), and discusses the community and the engineering project supporting the library.

By 2023, we will have implemented the open-source version of most general NLP capabilities available in English for Thai\footnote{\url{https://nlpforthai.com/}}. We see the following items as the next major milestones:
\begin{itemize}
    \item \textbf{Domain-specific datasets/models} Some capabilities are not performing well on specific use cases; for instances, named-entity recognition in financial reports, medical terms translation, and legal documents question answering. We believe more domain-specific datasets and models will help close this gap.
    \item \textbf{Robust benchmark for Thai NLP tasks} As NLP has garnered more attention, more models and datasets, both open- and closed-source, will be available. It will, therefore, be imperative to have a robust benchmark in comparing the models' performance and the datasets' quality.
    \item \textbf{Correctness and consistency} Search key generation (such as Soundex), sorting, and tokenization\footnote{Some phonetic algorithm and transliteration rely on syllable tokenization} have to be deterministic and strictly follow a specification, or an application may behave in an unexpected fashion. More test cases and verification might be needed for these features.
    \item \textbf{Efficient mechanism to load and manage datasets/models} To reduce the size of the library and to carter the use in a system with a restricted network connection\footnote{\url{https://github.com/PyThaiNLP/pythainlp/issues/298}}.
    \item \textbf{Seamless integration with language-agnostic tools} The ultimate goal is for developers to no longer need PyThaiNLP as Thai language is supported by standard NLP libraries such as spaCy and Hugging Face \cite{wolf-etal-2020-transformers}. We have begun this work with integrating our text processing functions and models to spaCy.
\end{itemize}

\section*{Acknowledgements}

First and foremost, we appreciate the contributions from all PyThaiNLP contributors\footnote{\url{https://github.com/PyThaiNLP/pythainlp/graphs/contributors}}. We would like to thank: 1) VISTEC-depa Thailand AI Research Institute and its director Sarana Nutanong for research collaboration and support in terms of academic guidance, computational resources, and personnel; 2) the companies featured in the industry impact section and respective interviewees Chrisada Sookdhis, Jayakorn Vongkulbhisal, Kowin Kulruchakorn, Phasathorn Suwansri, and Pongtachchai Panachaiboonpipop; 3) Ekapol Chuangsuwanich for academic guidance and contribution to models and datasets; 4) MacStadium for infrastructure support; and 5) NLP-OSS 2023 anonymous reviewers. We are much obliged to free and open-source software community for software building blocks and best practices, including but not limited to NumFOCUS, fast.ai, Hugging Face, and Thai Linux Working Group. Moreover, we thank organizations who care enough to develop multilingual resources to accommodate low-resource languages, most notably Meta AI. Lastly, we cannot thank enough volunteers of various open-content communities, including Wikipedia, Common Voice, TED Translators, and similar local initiatives; modern NLP will not be possible without their accumulated effort.

\section*{Limitations}

In our current CI workflow, every code commit to the repository triggers an automated test suit for all supported platforms. The process can be challenging if our package depends on large language models (LLMs) because a single LLM can exhaust the memory of our free-tier CI infrastructure. Some of the components can be cached to reduce build time, but they have to be loaded to the memory in any case. This forced us to drop some LLM-related tests and scarified the code coverage of the library as discussed in Section \ref{sec:software-quality}.

Even we have a resource to do such tests with the current design, it is neither economical nor sustainable. An improved test utilizing a stub, mock, or spy (proxy) test pattern that provides an off-line ``fake inference'' can help this. These techniques have been proven useful in other software testing involving expensive database/API queries or network connections. \citet{lyra_effective_2019} and \citet{microsoft_testing_2020} provide such examples, using the Python Standard Library's \texttt{unittest.mock}. This can reduce a number of times an LLM is actually being loaded/called. The required inference could be handled either by a non-free tier CI plan from the same or different provider (which should be more affordable now due to reduced number of calls) or by a computer outside the cloud.

\bibliography{custom}
\bibliographystyle{acl_natbib}

\end{document}